\documentclass[sigconf]{acmart}
\AtBeginDocument{%
  }

\copyrightyear{2026}
\acmYear{2026}
\setcopyright{cc}
\setcctype{by}
\acmConference[KDD '26]{Proceedings of the 32nd ACM SIGKDD Conference on Knowledge Discovery and Data Mining V.1}{August 09--13, 2026}{Jeju Island, Republic of Korea}
\acmBooktitle{Proceedings of the 32nd ACM SIGKDD Conference on Knowledge Discovery and Data Mining V.1 (KDD '26), August 09--13, 2026, Jeju Island, Republic of Korea}
\acmPrice{}
\acmDOI{10.1145/3770854.3785701}
\acmISBN{979-8-4007-2258-5/2026/08}
\acmJournal{JDS}
\acmVolume{37}
\acmNumber{4}
\acmArticle{111}
\acmMonth{8}

\definecolor{dark2green}{rgb}{0.1, 0.65, 0.3}
\definecolor{dark2orange}{rgb}{0.9, 0.4, 0.}
\definecolor{dark2purple}{rgb}{0.4, 0.4, 0.8}
\newcommand{\first}[1]{\textbf{\textcolor{dark2green}{#1}}}
\newcommand{\second}[1]{\textbf{\textcolor{dark2orange}{#1}}}
\newcommand{\third}[1]{\textbf{\textcolor{dark2purple}{#1}}}

\usepackage{cleveref}
\usepackage[utf8]{inputenc} % allow utf-8 input
\usepackage[T1]{fontenc}    % use 8-bit T1 fonts
\usepackage{hyperref}       % hyperlinks
\usepackage{url}            % simple URL typesetting
\usepackage{booktabs}       % professional-quality tables
\usepackage{amsfonts}       % blackboard math symbols
\usepackage{nicefrac}       % compact symbols for 1/2, etc.
\usepackage{microtype}      % microtypography
\usepackage{xcolor}         % colors
\usepackage{graphicx}
\usepackage{subfigure}
\usepackage{multirow}
\usepackage{wrapfig}

\begin{document}

\title{Benchmarking Positional Encodings for GNNs and Graph Transformers}

\author{Florian Grötschla}
\email{fgroetschla@ethz.ch}
\affiliation{%
  \institution{ETH Zurich}
  \city{Zurich}
  \country{Switzerland}
}

\author{Jiaqing Xie}
\email{jiaxie@ethz.ch}
\affiliation{%
  \institution{ETH Zurich}
  \city{Zurich}
  \country{Switzerland}
}

\author{Roger Wattenhofer}
\email{wattenhofer@ethz.ch}
\affiliation{%
  \institution{ETH Zurich}
  \city{Zurich}
  \country{Switzerland}
}

\renewcommand{\shortauthors}{Florian Grötschla, Jiaqing Xie, and Roger Wattenhofer}

\begin{abstract}
  Positional Encodings (PEs) are essential for injecting structural information into Graph Neural Networks (GNNs), particularly Graph Transformers, yet their empirical impact remains insufficiently understood. We introduce a unified benchmarking framework that decouples PEs from architectural choices, enabling a fair comparison across 8 GNN and Transformer models, 9 PEs, and 10 synthetic and real-world datasets. Across more than 500 model–PE–dataset configurations, we find that commonly used expressiveness proxies, including Weisfeiler–Lehman distinguishability, do not reliably predict downstream performance. In particular, highly expressive PEs frequently fail to improve, and can even degrade performance on real-world tasks. At the same time, we identify several simple and previously overlooked model–PE combinations that match or outperform recent state-of-the-art methods. Our results demonstrate the strong task-dependence of PEs and underscore the need for empirical validation beyond theoretical expressiveness. To support reproducible research, we release an open-source benchmarking framework for evaluating PEs for graph learning tasks.
\end{abstract}


%%
%% Article type: Research, Review, Discussion, Invited or position
\acmArticleType{Research}
%%
%% Links to code and data
\acmCodeLink{https://github.com/borisveytsman/acmart}
\acmDataLink{htps://zenodo.org/link}
%%
%% Authors' contribution
\acmContributions{BT and GKMT designed the study; LT, VB, and AP
  conducted the experiments, BR, HC, CP and JS analyzed the results,
  JPK developed analytical predictions, all authors participated in
  writing the manuscript.}
%%
%% Sometimes the addresses are too long to fit on the page.  In this
%% case uncomment the lines below and fill them accodingly.
%%
%% \authorsaddresses{Corresponding author: Ben Trovato,
%% \href{mailto:trovato@corporation.com}{trovato@corporation.com};
%% Institute for Clarity in Documentation, P.O. Box 1212, Dublin,
%% Ohio, USA, 43017-6221}

\maketitle

\section{Introduction}

\begin{figure*}
    \centering
    \includegraphics[width=0.95\linewidth]{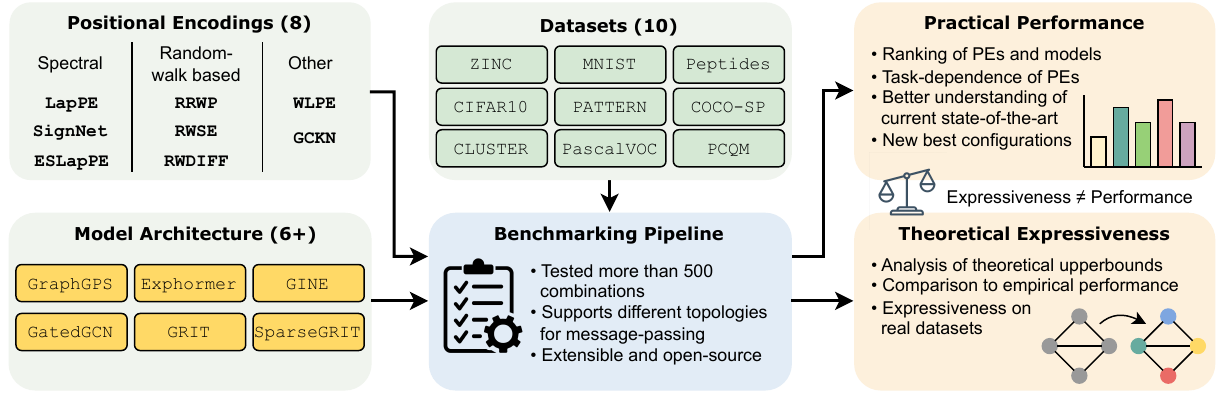}
    \caption{Conceptual overview of our benchmarking framework for PEs in GNNs and Graph Transformers. 
    Our empirical results show that practical performance does not always align with theoretical expressiveness, challenging conventional assumptions in the literature. 
    We identify new best-performing configurations by systematically exploring combinations across real and synthetic benchmarks on a wide variety of PEs, models and datasets.}
    \label{fig:overview}
\end{figure*}

Graphs are fundamental structures for modeling complex relationships in diverse domains, from social networks and molecular biology to recommendation systems. Graph Neural Networks (GNNs), particularly message-passing neural networks (MPNNs), have substantially advanced graph learning through their powerful ability to aggregate information from local neighborhoods. This local aggregation paradigm has led to substantial progress in node classification, link prediction, and graph regression tasks~\cite{kipf2016semi, wu2020comprehensive}. However, a fundamental limitation of MPNNs is their difficulty in capturing long-range dependencies, which are crucial in applications such as molecular interaction modeling and hierarchical social network analysis~\cite{xu2018powerful, alon2020bottleneck}.
To address these shortcomings, Graph Transformers (GTs) extend the self-attention framework to graphs, enabling global information exchange between all nodes~\cite{dwivedi2020generalization}. Unlike sequences in natural language processing, graphs lack a natural positional ordering, making Positional Encodings (PEs) necessary for embedding structural information~\cite{vaswani2017attention, black2024comparing}. These encodings provide geometric and topological information to otherwise position-agnostic neural architectures. Yet, incorporating effective PEs into graphs is considerably more difficult than in sequences because of the complex and non-linear structure of graph topology.
Although PEs play a central role in graph learning, their impact is often conflated with architectural innovations. Existing evaluations typically assess PEs within specific models, making it difficult to isolate their individual contributions. For example, PEs such as RWSE (Random Walk Structural Encoding) and LapPE (Laplacian Positional Encoding) have been evaluated primarily alongside specific architectures such as GraphGPS~\cite{rampavsek2022recipe} or Exphormer~\cite{shirzad2023exphormer}, rather than independently assessed across various models. Moreover, existing evaluations frequently rely on synthetic data or theoretical expressiveness metrics, such as Weisfeiler-Lehman (WL) distinguishability~\cite{xu2018powerful, sato2021random}, which often fail to correlate with practical performance on real-world tasks. This raises important questions regarding the utility of different PEs in various graph learning scenarios.

In this paper, we address these gaps through a comprehensive and systematic reassessment of Positional Encodings for GNNs. We propose a unified benchmarking framework explicitly designed to decouple the evaluation of PEs from architectural innovations. Our evaluation spans 8 graph architectures combined with 9 Positional Encodings across 10 synthetic and real-world datasets. In total, we analyze more than 500 model--PE--dataset configurations.
To the best of our knowledge, this constitutes one of the broadest empirical evaluations of PEs across both MPNNs and GTs under a unified protocol.
Our results reveal notable discrepancies between theoretical expressiveness and empirical performance. Positional Encodings such as RWSE exhibit strong theoretical capabilities on synthetic benchmarks yet frequently fail to translate these strengths into consistent performance improvements on real-world datasets. In contrast, spectral PEs demonstrate a robust balance between theoretical and empirical effectiveness and provide more reliable results across varied graph datasets. Further, we uncover several combinations of established architectures and PEs that match or outperform recent state-of-the-art methods on multiple benchmarks. These findings challenge the prevailing assumption that newer architectures or encodings inherently outperform established methods. Our benchmarking thus not only highlights overlooked configurations but also provides empirical insights that can help practitioners shortlist PEs for a given task and compute budget, while emphasizing the need for task-specific validation.
To facilitate reproducibility and continued evaluation in the community, we publicly release our benchmarking framework, including implementations of all PEs and model architectures tested in this study.
In summary, our main contributions are as follows.
\begin{itemize}
    \item We introduce a systematic benchmarking framework that explicitly isolates the evaluation of PEs from architectural innovations in GNNs and Graph Transformers.
    \item We empirically demonstrate that higher theoretical expressiveness does not reliably correlate with improved downstream performance across tasks and datasets.
    \item We identify multiple simple and previously underexplored combinations of established architectures and PEs that match or outperform recent state-of-the-art methods.
    \item We publicly release a reproducible and extensible open-source benchmark to support the evaluation of PEs in future work\footnote{\url{https://github.com/ETH-DISCO/Benchmarking-PEs}}.
\end{itemize}

\section{Related Work}

%\subsection{Positional Encodings}
\label{sec:positional_encodings}

\subsection{Graph Neural Network Architectures}

\paragraph{Message-Passing GNNs.}
Early architectures like GCN~\cite{kipf2016semi}, GraphSAGE~\cite{hamilton2017inductive}, GAT~\cite{velivckovic2017graph}, and GIN~\cite{xu2018powerful} established the message-passing framework, where nodes iteratively aggregate features from their neighbors. These models are typically bounded by the expressiveness of the Weisfeiler-Lehman (1-WL) test. Later variants improved this design: GatedGCN~\cite{bresson2017residual} adds gated residual connections, GINE~\cite{hu2019strategies} incorporates edge features, and PNA~\cite{corso2020principal} leverages multi-statistic aggregators. Others extended expressiveness using subgraph or motif counts~\cite{zhang2021nested} or operating on higher-order structures such as cell complexes~\cite{bodnar2021weisfeiler}. Still, most MPNNs struggle to capture long-range dependencies due to their inherently local nature.

\paragraph{Graph Transformers.}
Inspired by the transformer architecture \cite{vaswani2017attention}, Graph Transformers capture long-range dependencies via global self-attention mechanisms. Fully-connected variants can be powerful~\cite{kim2022pure}, but typically require graph-specific inductive biases~\cite{hussain2021edge}. Models like Graphormer~\cite{ying2021transformers} introduce shortest-path and centrality encodings directly into attention, while GraphGPS~\cite{rampavsek2022recipe} combines local MPNNs with global attention. Sparse attention models such as Exphormer~\cite{shirzad2023exphormer} improve scalability, and GRIT~\cite{ma2023graph} uses a more expressive attention mechanism with learnable edge updates. Transformers can integrate structural bias through learned masks~\cite{ma2023graph} and match MPNN expressiveness when equipped with strong Positional Encodings~\cite{black2024comparing} or virtual nodes~\cite{fu2024vcr}. In contrast, MPNNs can emulate Transformers under certain conditions~\cite{keriven2024functions}. 

\begin{table*}[t] \scriptsize
    \caption{Summary statistics of the datasets used in our benchmark, including graph size, connectivity, prediction level,
task type, and evaluation metric. It spans both small synthetic graphs and larger real-world benchmarks from different domains.}
    \centering
    \resizebox{\textwidth}{!}{%
    \begin{tabular}{lccccccl}
        \toprule
        \textbf{Dataset} & \textbf{\# Graphs} & \textbf{Avg.}  $|N|$  & \textbf{Avg.}  $|E|$ & \textbf{Directed} & \textbf{Prediction level} & \textbf{Prediction task} & \textbf{Metric} \\
        \midrule
        ZINC & 12,000 & 23.2 & 24.9 & No & graph & regression & Mean Abs. Error \\
        MNIST & 70,000 & 70.0 & 564.5 & Yes & graph & 10-class classif. & Accuracy \\
        CIFAR10 & 60,000 & 117.6 & 941.1 & Yes & graph & 10-class classif. & Accuracy \\
        PATTERN & 14,000 & 118.9 & 2,359.2 & No & inductive node & binary classif. & Accuracy \\
        CLUSTER & 12,000 & 117.2 & 1,510.9 & No & inductive node & 6-class classif. & Accuracy \\
        \midrule
         PascalVOC-SP & 11,355 & 479.4 & 2,710.5 & No & inductive node & 21-class classif. & F1 score \\
        COCO-SP & 123,286 & 476.4 & 2,693.7 & No & inductive node & 81-class classif. & F1 score \\
        PCQM-Contact & 529,434 & 30.1 & 69.1 & No & inductive link & link ranking & MRR (Fil.) \\
        Peptides-func & 15,535 & 150.9 & 307.3 & No & graph & 10-task classif. & Avg. Precision \\
        Peptides-struct & 15,535 & 150.9 & 307.3 & No & graph & 11-task regression & Mean Abs. Error \\
        \bottomrule
    \end{tabular}
    }
    \label{tab:dataset_stats}
\end{table*}

\subsection{Positional Encodings for Graphs}

Unlike grids or sequences, graphs lack a natural order or coordinate system. PEs inject structural information into node representations and are indispensable for Graph Transformers, which would otherwise treat nodes as a permutation-invariant set~\cite{dwivedi2020generalization, dwivedi2023benchmarking}. Even message-passing GNNs can benefit from PEs by receiving signals beyond the local neighborhood, thus improving their ability to capture long-range dependencies~\cite{dwivedi2021graph}. In general, PEs can be categorized into \textbf{spectral}, \textbf{random-walk}, and \textbf{other} types. While other taxonomies are possible, this grouping is useful empirically, as encodings within each category often exhibit similar behavior in our evaluation.

\paragraph{Spectral Encodings.}
Spectral PEs use eigenvectors of graph matrices (typically the Laplacian) as node coordinates. The Laplacian Positional Encoding (LapPE)~\cite{dwivedi2023benchmarking} embeds the nodes along the principal directions of the graph. SignNet~\cite{lim2022sign} addresses the sign and basis ambiguities of raw eigenvectors by learning invariant transformations. ESLapPE~\cite{wang2022equivariant} enhances stability under graph isomorphisms. GCKN~\cite{mialon2021graphit} approximates the spectral geometry through truncated kernel expansions. 

\paragraph{Random-Walk and Diffusion Encodings.}
RWSE~\cite{dwivedi2023benchmarking} encodes the probability of returning to the same node over multiple random walk lengths. Relative Random Walk PEs (RRWP), as used in GRIT~\cite{ma2023graph}, compute stationary distributions under personalized PageRank (PPR)-like schemes~\cite{gasteiger2018predict}. Related diffusion methods, such as heat kernels and resistance distances, have also been explored~\cite{dong2019learning}. These encodings capture local and global connectivity, but can be computationally expensive for large graphs~\cite{tonshoff2023did}.

\paragraph{Others.}
Distance Encoding (DE)~\cite{li2020distance} enhances GNN expressiveness by adding distances to reference nodes. Graphormer~\cite{ying2021transformers} applies the shortest-path distances and edge connectivity as relative PE in the attention matrices. Node degrees, centrality, and WL-labels (WLPE)~\cite{dwivedi2020generalization} also encode useful structural signals. Subgraph-based encodings~\cite{zhao2021stars} extend PEs with local structure indicators beyond the 1-hop neighborhood. These features can be used in either absolute or relative form, and are especially effective in Transformer-based models.

\subsection{Benchmarking of PEs}

Recent work has shown that improvements in graph models often stem from either the architecture or the PE, or both~\cite{black2024comparing, keriven2024functions}. For example, performance gains attributed to architectural changes in Transformers have sometimes been primarily due to the choice of PE. To fairly assess their respective contributions, it is crucial to decouple these components.
\citet{black2024comparing} provide a theoretical comparison of Graph Transformers with different PEs, demonstrating that certain encodings can render models equivalent in expressive power. \citet{keriven2024functions} further analyze how the functional capacity of GNNs depends on the structure of their PEs. These studies suggest that sufficiently powerful PEs can enable even 1-WL-limited GNNs to solve more complex tasks, provided the model can make use of them.
Despite these insights, a practical benchmark for evaluating diverse combinations of PEs and architectures was lacking. Our work fills this gap by systematically evaluating a wide range of PEs (spectral, random-walk, structural) across both message-passing and Transformer-based models. By controlling for confounding factors, our benchmark isolates the effect of PEs across tasks and architectures. Our benchmark shows that, while the effectiveness of PE varies by task, certain encodings provide consistent gains across multiple architectures. This supports the view that architectures and PEs can often be composed modularly, allowing strong components from both domains to be combined. Thus, our benchmark provides guidance for selecting or designing PEs in practice.

\section{Topology and Attention}
\label{sec:bridging_gts_wl}

The effectiveness of PEs depends not only on their formulation, but also on how they interact with the underlying model architecture. In particular, the topology over which information is exchanged, whether local neighborhoods in message-passing GNNs or fully connected attention in GTs, can influence the extent to which a PE contributes to performance. %This section analyzes how architectural connectivity affects the empirical utility of PEs.
For GTs, full self-attention allows each node to attend to all others and naturally allows direct global information exchange. Although this design can compensate for the absence of long-range structural features in the input, it also reduces the reliance on PEs to encode this information. However, full attention comes with substantial computational overhead, particularly on large, sparse graphs. This raises a practical question: \emph{To what extent is full attention necessary for competitive performance and does the answer depend on the presence of PEs?}
To study this, we introduce a sparsified variant of the GRIT architecture~\citep{ma2023graph}, referred to as \emph{Sparse GRIT}. It retains GRIT’s attention mechanism and edge update scheme, but restricts attention to a node's original neighbors, rather than using a fully connected topology. This effectively transforms the GT into a sparse, local message-passing network, allowing us to isolate the impact of the update mechanism. To complement the analysis, we also evaluate message-passing convolutions on fully-connected graphs to test whether an attention mechanism is necessary to perform well on fully-connected graphs. 
This perspective allows us to view the considered architectures within a unified message-passing framework. In the case of (fully-connected) GTs, we only need to change the underlying message-passing topology to a fully-connected graph~\cite{velivckovic2023everything}. 

\paragraph{Sparse GRIT Convolution.}
Sparse GRIT applies the same edge update rules as GRIT~\cite{ma2023graph}, using updated edge encodings $\hat{\mathbf{e}}_{i,j}$, but only on the edges present in the input graph. This distinguishes it from fully connected attention mechanisms and also from local attention mechanisms like GAT, which do not update edge representations. The node update rule is given by (adapted from \citet{ma2023graph}):
\begin{align*}
\hat{\mathbf{x}}_i = \sum_{j \in \mathcal{N}(i)} 
\frac{e^{w_j \cdot \hat{\mathbf{e}}_{i,j} }}{\sum_{k \in \mathcal{N}(i)}  e^{w_k \cdot \hat{\mathbf{e}}_{i,k}} } \cdot \left( \mathbf{W}_V \mathbf{x}_j + \mathbf{W}_{E_V} \hat{\mathbf{e}}_{i,j} \right),
\end{align*}
where $w_j$ denotes the attention weight, and $\mathbf{W}_V$, $\mathbf{W}_{E_V}$ are learnable projection matrices. The key distinction from GRIT is that attention is computed only over local neighborhoods $\mathcal{N}(i)$, using a sparse softmax normalization.
This design allows Sparse GRIT to scale more efficiently while preserving the core inductive biases of GRIT. Importantly, it enables us to evaluate how Positional Encodings perform under different connectivity regimes. As we show in Section~\ref{sec:eval}, Sparse GRIT often matches the performance of GRIT, despite using significantly fewer edges. This suggests that full attention may not be required in settings where local structure is predictive and that PEs become increasingly important in sparser architectures where less global information is available through the model itself.

\section{Benchmarking Framework}
\label{sec:benchmarking_framework}

We perform a benchmarking of state-of-the-art models combined with commonly used PEs to identify optimal configurations. This analysis addresses a common gap in the literature, where new PEs are introduced alongside novel architectures but are rarely evaluated independently of existing models. %By decoupling architectures from PEs, our approach enables a comprehensive exploration of possible combinations.
To enable the evaluation of models and future research for measuring the impact of Positional Encodings, we provide a unified codebase that includes the implementation of all tested models and the respective Positional Encodings. We base the code on GraphGPS~\cite{rampavsek2022recipe} and integrate all missing implementations. This makes for reproducible results and easy extensibility for new datasets, models, or PEs. Our codebase also provides readily available implementations for NodeFormer \citep{wu2022nodeformer}, Difformer \citep{wu2023difformer}, GOAT \citep{kong2023goat}, GraphTrans \citep{wu2021representing}, GraphiT \citep{mialon2021graphit}, and SAT \citep{chen2022structure} that are based on the respective codebases.

In our experiments, we used five different random seeds for the \textsc{BenchmarkingGNNs} datasets~\citep{dwivedi2023benchmarking} and four for the others. 
All experiments can be executed on a single Nvidia RTX 3090 (24GB) or a single RTX A6000 (40GB).
To avoid out-of-memory (OOM) issues on LRGB datasets, we reserve up to 100GB of host memory for preprocessing Positional Encodings.
For configurations that exceeded this compute/memory envelope (e.g., due to PE preprocessing or attention quadratic costs), we mark them as infeasible and exclude them consistently across models. The codebase can be found online: \url{https://github.com/ETH-DISCO/Benchmarking-PEs}.

\subsection{Datasets and Model Configurations} 
\label{sec:datasets}

We begin by describing the datasets and model configurations used in our benchmark, which define the fixed experimental setting under which all Positional Encodings are evaluated.

\textbf{BenchmarkingGNNs} includes \textit{MNIST}, \textit{CIFAR10}, \textit{CLUSTER}, \textit{PATTERN}, and \textit{ZINC}, following the protocols established in \textit{GraphGPS} \citep{rampavsek2022recipe}, \textit{Exphormer} \citep{shirzad2023exphormer}, and \textit{GRIT} \citep{ma2023graph}. These datasets have traditionally been used to benchmark Graph Neural Networks (GNNs) \citep{dwivedi2023benchmarking}, excluding Graph Transformers. 
We adhere to established settings from the relevant literature for each model. Specifically, for GatedGCN and GraphGPS, we follow the configurations detailed for GraphGPS~\citep{rampavsek2022recipe}. For Exphormer, we utilize the settings from the original paper~ \citep{shirzad2023exphormer}. For GINE, Sparse GRIT, and global GRIT models, we adopt the configurations from GRIT~\citep{ma2023graph}.

\textbf{Long-Range Graph Benchmark} (LRGB) \citep{dwivedi2022long} includes \textit{Peptides-func}, \textit{Peptides-struct}, \textit{PascalVOC-SP}, \textit{PCQM-Contact}, and \textit{COCO}. The tasks have been developed to necessitate long range interactions. We consider four models: GatedGCN, GraphGPS, Exphormer, and Sparse GRIT. For GatedGCN and GraphGPS, we mainly follow the fine-tuned configurations as described by \citet{tonshoff2023did}. For sparse GRIT, we adopt the hyperparameters used for the Peptides-func and Peptides-struct datasets and transfer these settings to COCO-SP, Pascal-VOC, and PCQM-Contact, as detailed by \citet{dwivedi2022long}. For Exphormer, we follow the configurations proposed by \citet{shirzad2023exphormer}.

Table~\ref{tab:pe_robustness} shows that, while absolute performance varies under
equal-budget tuning, the relative ordering of Positional Encodings is stable
across the depth and dropout settings tested. This ablation is not exhaustive:
a full per-configuration hyperparameter search over $>500$ combinations would be
computationally infeasible and is not standard practice for these datasets~\cite{rampavsek2022recipe, dwivedi2022long}. Instead, these controlled checks provide evidence that our main
ranking trends are robust to non-trivial hyperparameter variation.

\begin{table}[b]
\centering
\small
\caption{Robustness of Positional Encoding rankings under controlled equal-budget tuning.
We vary model depth ($L$) and dropout ($D$) for representative settings.
CIFAR10 is evaluated using accuracy ($\uparrow$), while ZINC is evaluated using mean absolute error (MAE, $\downarrow$).
Absolute performance changes, but relative PE rankings remain stable.}
\label{tab:pe_robustness}
\resizebox{\linewidth}{!}{%
\begin{tabular}{lc|ccc}
\toprule
Dataset / Model & Var. & LapPE & RWSE & SignNet \\
\midrule
\multirow{6}{*}{CIFAR10 / GRIT}
 & $L=3$   & 73.33$\pm$0.51 & 73.65$\pm$0.62 & 72.81$\pm$0.48 \\
 & $L=5$   & 74.45$\pm$0.28 & 75.28$\pm$0.43 & 73.97$\pm$0.05 \\
 & $L=7$   & 76.03$\pm$0.31 & 76.45$\pm$0.16 & 75.86$\pm$0.54 \\
\cmidrule(lr){2-5}
 & $D=0.1$ & 74.03$\pm$1.17 & 74.05$\pm$0.11 & 73.34$\pm$0.38 \\
 & $D=0.3$ & 73.48$\pm$1.04 & 73.30$\pm$0.79 & 73.26$\pm$0.19 \\
 & $D=0.5$ & 73.33$\pm$0.51 & 73.65$\pm$0.62 & 72.81$\pm$0.48 \\
\midrule
\multirow{6}{*}{ZINC / GatedGCN}
 & $L=4$   & 0.172$\pm$0.002 & 0.102$\pm$0.003 & 0.106$\pm$0.002 \\
 & $L=6$   & 0.155$\pm$0.004 & 0.103$\pm$0.007 & 0.106$\pm$0.002 \\
 & $L=8$   & 0.167$\pm$0.006 & 0.116$\pm$0.006 & 0.110$\pm$0.002 \\
\cmidrule(lr){2-5}
 & $D=0.0$ & 0.172$\pm$0.002 & 0.102$\pm$0.003 & 0.106$\pm$0.002 \\
 & $D=0.3$ & 0.192$\pm$0.010 & 0.135$\pm$0.050 & 0.147$\pm$0.000 \\
 & $D=0.5$ & 0.260$\pm$0.005 & 0.205$\pm$0.009 & 0.218$\pm$0.021 \\
\bottomrule
\end{tabular}
}
\end{table}

\paragraph{Scalability to Larger Graphs.}
We additionally evaluated Positional Encoding preprocessing on
large-scale graphs such as \textsc{ogbn-products} and \textsc{ogbn-mag}.
In these settings, only no-PE and lightweight encodings such as WLPE
are feasible. Spectral and random-walk-based encodings (e.g., LapPE,
RWSE) exceed available memory due to dense matrix operations.
As our goal is a controlled comparison of the effectiveness of PE, such
datasets are outside the feasible scope of our benchmark, at least with the PEs we evaluate. Implementations and configuration files for large-scale graphs are available in the codebase.
Statistics and prediction tasks for the datasets used are listed in Table \ref{tab:dataset_stats}.

\begin{table*}[t] \scriptsize
\centering
\caption{Results for the best-performing models and the PE they use for the \textsc{BenchmarkingGNNs} datasets. All runs except those for EGT and TIGT were done by us. SparseGRIT performs on par with GRIT on most datasets, indicating that, on the datasets we evaluate, full attention is not always
necessary to achieve competitive performance. We color the \first{best}, \second{second best}, and \third{third best} models.}
\begingroup
\setlength{\tabcolsep}{2pt}

\resizebox{\textwidth}{!}{%
\begin{tabular}{lclclclclcl}
\toprule
Model & CIFAR10 $\uparrow$ &  & CLUSTER $\uparrow$ &  & MNIST $\uparrow$ &  & PATTERN $\uparrow$ &  & ZINC $\downarrow$ &  \\
\midrule
EGT~\citep{hussain2022global} & $68.70 \pm 0.41$ & & \third{$\mathbf{79.23 \pm 0.35}$} & & $98.17 \pm 0.09$ & & $86.82 \pm 0.02$ & & $0.108 \pm 0.009$ & \\
TIGT~\citep{choi2024topology} & $73.96 \pm 0.36$ & & $78.03 \pm 0.22$ & & \second{$\mathbf{98.23 \pm 0.13}$} & & $86.68 \pm 0.06$ & & \first{$\mathbf{0.057 \pm 0.002}$} & \\
\midrule
GINE & $66.14 \pm 0.31$ & \tiny{(ESLapPE)} & $59.66 \pm 0.63$ & \tiny{(SignNet)} & $97.75 \pm 0.10$ & \tiny{(RWDIFF)} & $86.69 \pm 0.08$ & \tiny{(RWSE)} & $0.075 \pm 0.006$ & \tiny{(RWDIFF)} \\
GatedGCN & $69.57 \pm 0.79$ & \tiny{(RRWP)} & $75.29 \pm 0.05$ & \tiny{(SignNet)} & $97.91 \pm 0.08$ & \tiny{(RRWP)} & $86.83 \pm 0.03$ & \tiny{(RWSE)} & $0.102 \pm 0.003$ & \tiny{(RWSE)} \\
SparseGRIT & \third{$\mathbf{74.95 \pm 0.26}$} & \tiny{(RRWP)} &  \first{$\mathbf{79.87 \pm 0.08}$} & \tiny{(RRWP)} & $98.12 \pm 0.05$ & \tiny{(RWSE)} & \second{$\mathbf{87.17 \pm 0.04}$} & \tiny{(RRWP)} & \third{$\mathbf{0.065 \pm 0.003}$} & \tiny{(RRWP)} \\
Exphormer & \second{$\mathbf{75.21 \pm 0.10}$} & \tiny{(LapPE)} & $78.28 \pm 0.21$ & \tiny{(SignNet)} & \first{$\mathbf{98.42 \pm 0.18}$} & \tiny{(RRWP)} & $86.82 \pm 0.04$ & \tiny{(RWSE)} & $0.092 \pm 0.007$ & \tiny{(SignNet)} \\
\midrule
GRIT & \first{$\mathbf{75.66 \pm 0.41}$} & \tiny{(RRWP)} & \second{$\mathbf{79.81 \pm 0.11}$} & \tiny{(RRWP)} & $98.12 \pm 0.14$ & \tiny{(RRWP)} & \first{$\mathbf{87.22 \pm 0.03}$} & \tiny{(RRWP)} & \second{$\mathbf{0.059 \pm 0.001}$} & \tiny{(RRWP)} \\
GatedGCN (FC) & $71.08 \pm 0.60$ & \tiny{(RRWP)} & $74.78 \pm 0.46$ & \tiny{(SignNet)} & \third{$\mathbf{98.20 \pm 0.15}$} & \tiny{(GCKN)} & $86.85 \pm 0.02$ & \tiny{(RWSE)} & $0.114 \pm 0.003$ & \tiny{(RWSE)} \\
GraphGPS & $72.31 \pm 0.20$ & \tiny{(noPE)} & $78.31 \pm 0.11$ & \tiny{(SignNet)} & $98.18 \pm 0.12$ & \tiny{(ESLapPE)} & \third{$\mathbf{86.87 \pm 0.01}$} & \tiny{(RWSE)} & $0.074 \pm 0.006$ & \tiny{(RWSE)} \\
\bottomrule
\end{tabular}
}
\endgroup
\label{tab:sota}
\end{table*}

\begin{figure*}[t]
    \centering
    \includegraphics[width=\linewidth]{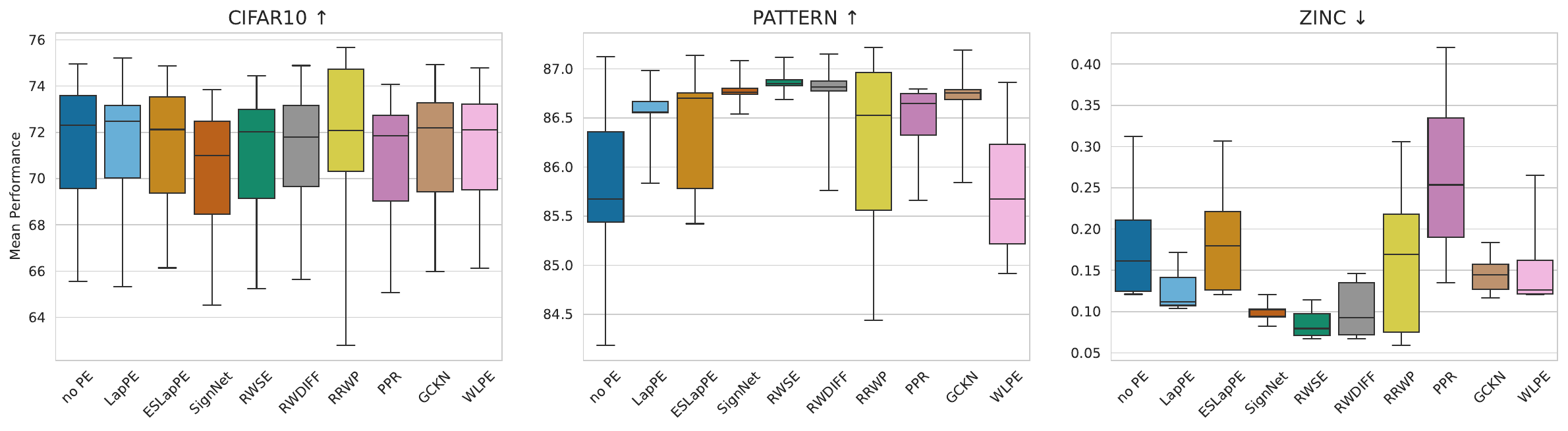}
    \caption{Performance comparison of target metrics across selected datasets from \textsc{BenchmarkingGNNs}. The boxplots illustrate the performance range for all models included in the study, with whiskers representing the minimum and maximum performance observed. Notably, RRWP consistently achieves the best results, whereas certain PEs, such as SignNet on CIFAR10, can sometimes decrease performance relative to the baseline without PEs. %Plots for the remaining datasets are provided in Appendix~\ref{sec:complete_results}.
    }
    \label{fig:boxplot_benchmark}
\end{figure*}

\subsection{WL Distinguishability}
\label{sec:wl_distinguishability}

Several parts of our analysis rely on \emph{WL distinguishability} as a proxy
for the structural expressiveness induced by a PE.
We therefore formally define this notion here.

\begin{figure}
    \centering
    \includegraphics[width=\linewidth]{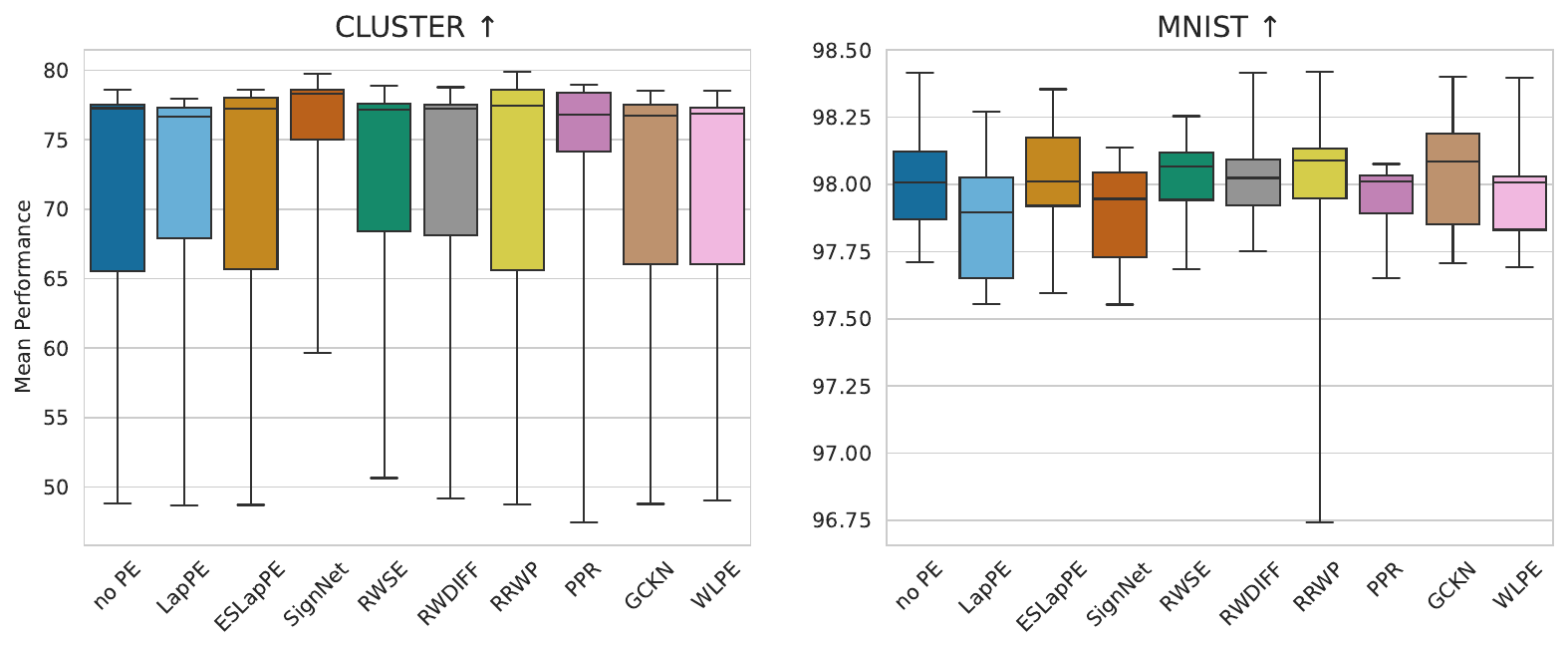}
    \caption{Mean performance of different Positional Encodings on more datasets from the \textsc{BenchmarkingGNNs}.}
    \label{fig:improvement_plot_complete_mean_gb}
\end{figure}

\begin{table*}[t] \scriptsize
\caption{Best-performing models and PEs for the LRGB datasets under the LRGB evaluation protocol. On PCQM-Contact, Exphormer+LapPE achieves the best performance among the methods evaluated in our benchmark.}
\centering

\begingroup
\setlength{\tabcolsep}{2pt}

\resizebox{\linewidth}{!}{%
\begin{tabular}{lclclclclcl}
\toprule

Model  & COCO-SP $\uparrow$&  & PCQM-Contact $\uparrow$&  & PascalVOC-SP $\uparrow$&  & Peptides-func $\uparrow$&  & Peptides-struct $\downarrow$&  \\
\midrule
GCN~\citep{tonshoff2023did} & $13.38 \pm 0.07$ & & $45.26 \pm 0.06$ & & $0.78 \pm 0.31$ & & $68.60 \pm 0.50$ & & $24.60 \pm 0.07$  & \\
GINE~\citep{tonshoff2023did} & $21.25 \pm 0.09$ & & $46.17 \pm 0.05$ & & $27.18 \pm 0.54$ & & $66.21 \pm 0.67$ & & $24.73 \pm 0.17$ & \\
GatedGCN~\citep{tonshoff2023did} & $29.22 \pm 0.18$ & & \third{$\mathbf{46.70 \pm 0.04}$} & & $38.80 \pm 0.40$ & & $67.65 \pm 0.47$ & & $24.77 \pm 0.09$ & \\
CRaWl~\citep{tonshoff2021walking} & - & & - & & \first{$\mathbf{45.88 \pm 0.79}$}  & & \third{$\mathbf{70.74 \pm 0.32}$} & & $25.06 \pm 0.22$ & \\
S$^2$GCN~\citep{geisler2024spatio} & - & & - & & - & & \first{$\mathbf{73.11 \pm 0.66}$} & & \first{$\mathbf{24.47 \pm 0.32}$} & \\
DRew~\citep{gutteridge2023drew} & - & & $34.42 \pm 0.06$ &  & $33.14 \pm 0.24$ &  & \second{$\mathbf{71.50 \pm 0.44}$} &  & $25.36 \pm 0.15$ & \\
Graph ViT~\citep{he2023generalization} & - & & - & & - & & $68.76 \pm 0.59$ & & \third{$\mathbf{24.55 \pm 0.27}$} & \\
GatedGCN-VN~\citep{rosenbluth2024distinguished} & \third{$\mathbf{32.44 \pm 0.25}$} &  & - & & \third{$\mathbf{44.77 \pm 1.37}$} & & $68.23 \pm 0.69$ & & $24.75 \pm 0.18$ & \\
\midrule
Exphormer & \second{$\mathbf{34.85 \pm 0.11}$} & \tiny{(ESLapPE)} & \first{$\mathbf{47.37 \pm 0.24}$} & \tiny{(LapPE)} & $42.42 \pm 0.44$ & \tiny{(LapPE)} & $64.24 \pm 0.63$ & \tiny{(LapPE)} & $24.96 \pm 0.13$ & \tiny{(LapPE)} \\
%GatedGCN & $0.6453 \pm 0.0059$ & \tiny{(LapPE)} & $0.3633 \pm 0.0041$ & \tiny{(WLPE)} & - &  & - &  & - &  \\
GraphGPS & \first{$\mathbf{38.91 \pm 0.33}$} & \tiny{(RWSE)} & \second{$\mathbf{46.96 \pm 0.17}$} & \tiny{(LapPE)} & \second{$\mathbf{45.38 \pm 0.83}$} & \tiny{(ESLapPE)} & $66.20 \pm 0.73$ & \tiny{(LapPE)} & $24.97 \pm 0.24$ & \tiny{(LapPE)} \\
SparseGRIT & $19.76 \pm 0.38$ & \tiny{(noPE)} & $45.85 \pm 0.11$ & \tiny{(LapPE)} & $35.19 \pm 0.40$ & 
\tiny{(GCKN)} & $67.02 \pm 0.80$ & \tiny{(RRWP)} & $24.87 \pm 0.14$ & \tiny{(LapPE)} \\
GRIT & $21.28 \pm 0.08$ & \tiny{(RWDIFF)} & $46.08 \pm 0.07$ & \tiny{(SignNet)}   & $35.56 \pm 0.19$ &  \tiny{(noPE)} & $68.65 \pm 0.50$ & \tiny{(RRWP)} & \second{$\mathbf{24.54 \pm 0.10}$} & \tiny{(RRWP)}   \\
\bottomrule
\end{tabular}%
}
\endgroup

\label{tab:sota_2}
\end{table*}

\begin{figure*}
    \centering
    \includegraphics[width=\linewidth]{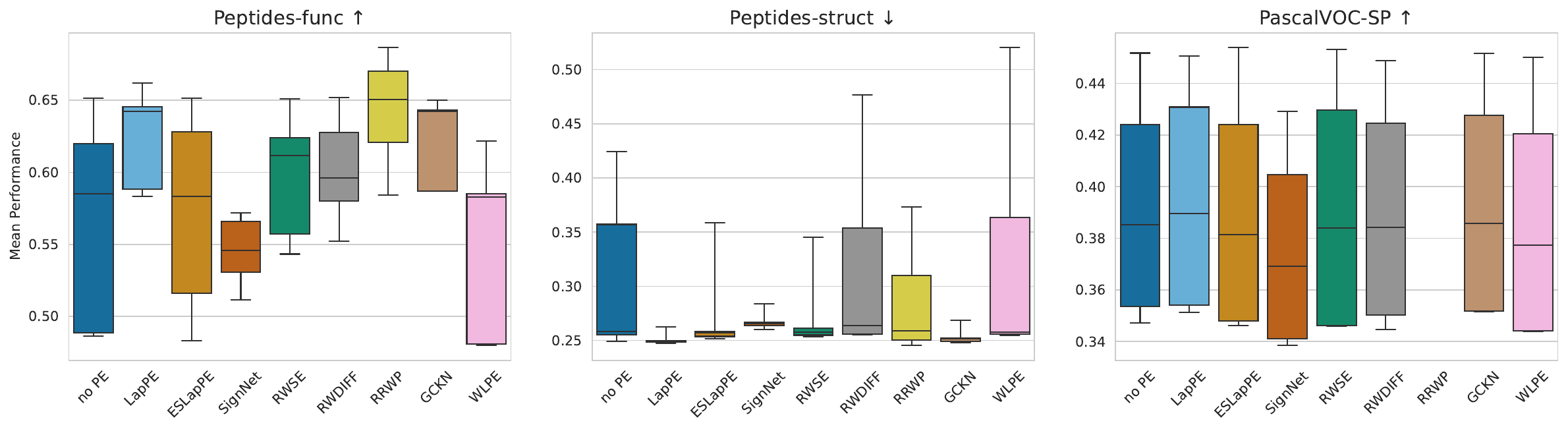}
    \caption{Performance comparison of target metrics across selected datasets from the Long-Range Graph Benchmark. The boxplots illustrate the performance range of all models included in the study, with whiskers indicating the minimum and maximum performance observed.  Plots for the remaining datasets are provided in Figure~\ref{fig:improvement_plot_complete_mean_lrgb}.}
    \label{fig:boxplot_lrgb}
\end{figure*}

\begin{figure}[t]
    \centering
    \includegraphics[width=\linewidth]{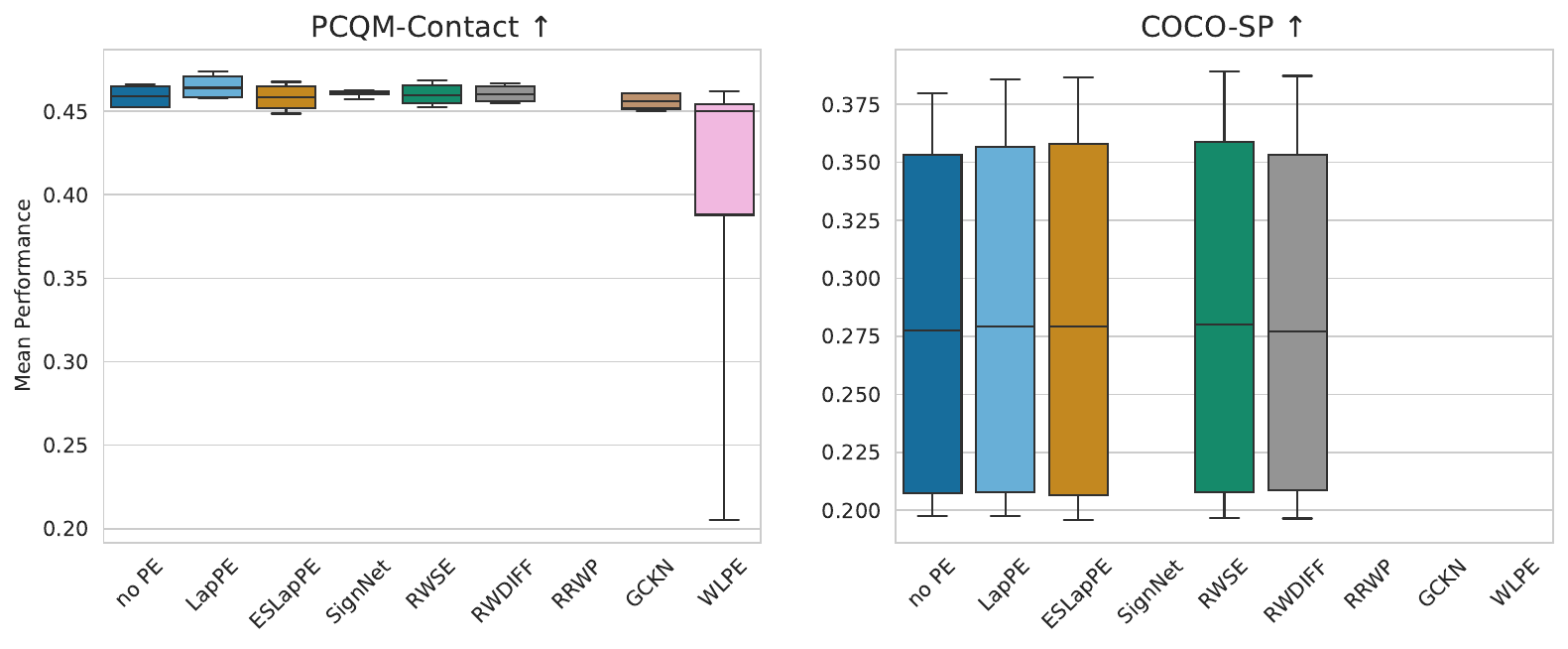}
    \caption{Mean performance of different Positional Encodings on more datasets from the Long Range Graph Benchmark.}
    \label{fig:improvement_plot_complete_mean_lrgb}
\end{figure}

Given a graph $G = (V,E)$ with initial node features (including positional 
encoding), we run $r$ rounds of the 1-dimensional Weisfeiler--Lehman (1-WL)
algorithm. Let $p$ denote the number of distinct node color classes obtained
after $r$ refinement rounds. We define the WL distinguishability score as
\begin{equation}
    \mathrm{WL\text{-}dist}(G) \;=\; \frac{p}{|V|} \in (0,1].
\end{equation}
A value of $1$ indicates that all nodes are structurally distinguishable under
1-WL, while smaller values indicate that nodes cannot be uniquely identified.
For example, on a cycle graph with identical initial node features, 1-WL assigns all nodes the same color in every round,
so $p=1$ and the score equals $1/|V|$ (minimal distinguishability).

\paragraph{Interpretation and Limitations.}
WL distinguishability provides an upper bound on the structural discrimination
capacity that a message-passing GNN could theoretically exploit when equipped
with a given PE. However, it is agnostic to task semantics, noise, and feature
distributions. As a result, higher WL distinguishability does not reliably
translate into improved downstream performance. In some cases, highly expressive
PEs may introduce inductive biases that are misaligned with the task, leading to
overfitting or degraded performance.

\begin{figure*}[t]
    \centering
    \includegraphics[width=\linewidth]{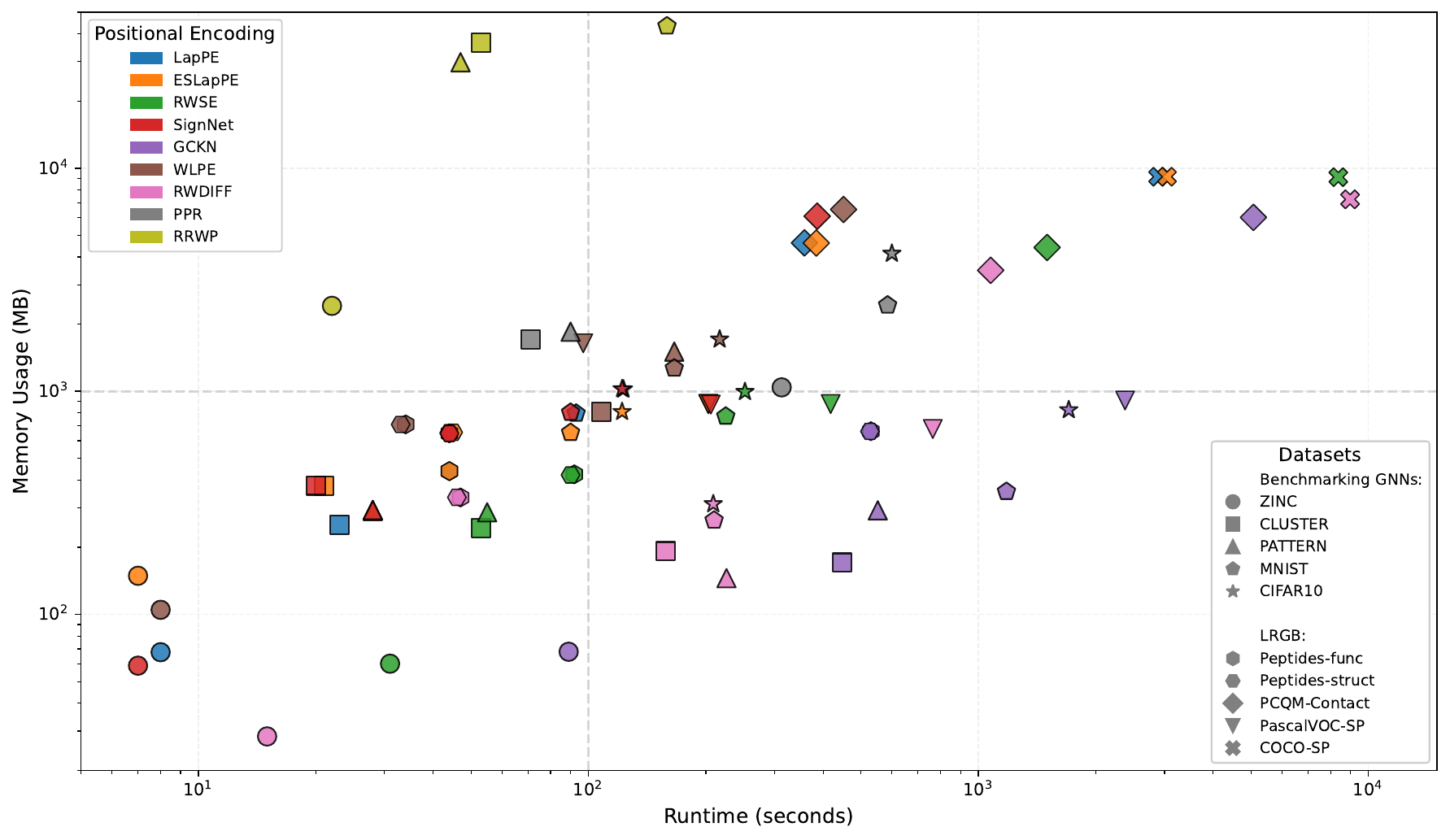}
    \caption{
    Computational requirements of Positional Encodings across benchmark datasets. We compare runtime versus memory usage. LapPE and ESLapPE are amongst the most efficient. In contrast, RRWP exhibits the highest resource consumption, with memory usage exceeding 40,000 MB on several datasets and failing to compute (OOM) on larger instances. SignNet, GCKN, and WLPE also encounter memory limitations on the COCO-SP dataset.
    }
    \label{fig:runtime_memory_pe_main}
\end{figure*}

\section{Performance Comparison}
\label{sec:eval}
Based on the framework we established in Section~\ref{sec:benchmarking_framework}, we benchmark the performance of different PEs on the \textsc{BenchmarkingGNNs}~\citep{dwivedi2023benchmarking} and \textsc{LRGB}~\citep{dwivedi2022long} datasets. Throughout this section, we emphasize relative ranking trends rather than absolute performance values, as our goal is to compare Positional Encodings under standardized settings rather than to optimize individual configurations.

\paragraph{Benchmarking GNNs Datasets.}

We first conduct a dataset-centric analysis where we assess the impact of various PEs on model performance. Figure~\ref{fig:boxplot_benchmark} presents the range of target metric values achieved across different PEs, aggregated over all models. Unaggregated results are provided in Table~\ref{GNNbenchmark} and \ref{LRGB} of the Appendix. Although most prior results were reproducible, we consistently observed slightly lower values for GRIT, even when using its official codebase and configurations.
Our findings reveal that PEs can significantly influence model performance, with the best choice of PE varying depending on the dataset and task. However, PEs can also negatively impact performance in some cases. For example, while RRWP performs best on the CIFAR10 dataset and ZINC, there are not always clear winners. Sometimes, good performance can be achieved even without any Positional Encoding (e.g., for PATTERN).
This is also evident when examining the best-performing configurations for each model and PE. The full set of runs (all model--PE--dataset configurations),
including configuration files and raw logs, is available in our released benchmarking codebase.
We summarize the best-performing configurations for the \textsc{BenchmarkingGNNs} datasets in Table~\ref{tab:sota}, where we can observe which PE led to the best performance for each model and dataset. This enables a fair comparison of all architectures and helps to determine which PE works best.
In our comparison, we observe that the sparse GRIT convolution emerges as the best graph convolution for sparse topologies. It competes effectively with full GRIT attention in most datasets. This suggests that these datasets do not require extensive long-range information exchange and can achieve strong performance with sparse message-passing. The GatedGCN convolution on the fully-connected graph does perform better than the original overall, but generally lags behind attention-based layers. Regarding the effectiveness of different PEs, random-walk-based encodings such as RRWP and RWSE consistently perform well in all tested models. The only notable exception is the CLUSTER dataset, where SignNet performs competitively for some architectures, although the best results are still achieved with RRWP.

\paragraph{Long-Range Graph Benchmark.}
\label{sec:lrgb_eval}

We extend our evaluation to the LRGB datasets and use hyperparameter configurations based on those of~\citet{tonshoff2023did}, with results presented in Table~\ref{tab:sota_2}. In these datasets, Laplacian-based encodings generally outperform others (except for the Peptides variations), likely due to their ability to capture more global structure in the slightly larger graphs.
This might also be reflected in the fact that transformer-based architectures or models that facilitate global information exchange consistently perform better. Our findings largely align with previous rankings, except for PCQM-Contact, where Exphormer achieves the best performance among the methods evaluated in our benchmark, which underscores the importance of thorough benchmarking of existing models.

Figure~\ref{fig:boxplot_lrgb} further analyzes the performance of the employed PEs. It should be noted that RRWP could not be utilized for larger datasets due to its significant memory footprint and computational complexity, similar to models employing full attention mechanisms. The results align with our previous analysis and show that on datasets like Peptides-func, the PE has a consistent impact on the performance, even when the values are aggregated over different architectures. This impact can also be negative compared to the baseline that does not use any PE. On other datasets (for example, PascalVOC-SP), the PE seems to play a lesser role, and good results can be achieved without any PE.

\subsection{Preprocessing Cost of Positional Encodings}
\label{sec:costs}

Although much of the existing literature focuses on the expressiveness and accuracy impact of PEs, practical adoption also depends on their computational overhead. Some PEs require solving large linear systems, computing high-order spectral decompositions, or simulating random walks, all of which can become prohibitive on large-scale graphs.
To quantify this overhead, we benchmark the preprocessing cost of each PE in terms of both runtime and peak memory usage. We report measurements on four representative datasets that differ in size and structure: \textsc{ZINC}, \textsc{CLUSTER}, \textsc{PCQM-Contact}, and \textsc{COCO-SP}. This subset includes both molecule-level graphs and large image-derived superpixel graphs.
\Cref{fig:runtime_memory_pe_main} summarizes the results. Among the evaluated PEs, LapPE, ESLapPE, and SignNet show consistently low preprocessing time and memory usage. In contrast, GCKN, RWSE, and RWDIFF incur significantly higher costs, particularly on larger graphs. For example, on \textsc{COCO-SP}, both SignNet and GCKN exceeded GPU memory limits during preprocessing and could not be run.
This analysis highlights an important trade-off: while more expressive PEs may offer potential performance benefits, their computational cost can limit practical applicability in real-world systems, particularly on high-resolution graphs or large datasets. As such, empirical performance should be evaluated not only in terms of accuracy, but also in light of resource constraints and scalability.

\section{The Impact of Expressiveness}
\label{sec:expressiveness}

Prior work has primarily evaluated PEs through theoretical lenses such as WL distinguishability~\cite{keriven2024functions, black2024comparing}. While such metrics capture structural expressiveness, their practical relevance remains uncertain. We therefore investigate whether higher theoretical expressiveness correlates with downstream performance, using both synthetic expressiveness datasets and real-world benchmarks.

%\subsection{Synthetic Expressiveness Datasets}
\label{sec:synth_eval}

\begin{table}[t]
    \centering
        \caption{WL distinguishability scores for PEs under sparse and fully-connected topologies on synthetic datasets. RWSE and LapPE consistently achieve perfect or near-perfect scores. Higher expressiveness under idealized conditions does not necessarily translate to better downstream performance.}
        \resizebox{\linewidth}{!}{%
    \begin{tabular}{llccccc}
\toprule
PE & Topology & \textsc{LIMITS 1} & \textsc{LIMITS 2} & \textsc{SKIP-CIRCLES} & \textsc{TRIANGLES} & \textsc{4-CYCLES}\\
 \midrule
\multirow{2}{*}{no PE} & sparse & 0.5 & 0.5 & 0.1 & 0.65 & 0.5 \\
 & fully-connected & 0.5 & 0.5 & 0.1 & 0.65 & 0.69 \\
\midrule
\multirow{2}{*}{GCKN} & sparse & 1.0 & 1.0 & 0.35 & 0.65 & 0.94 \\
 & fully-connected & 1.0 & 1.0 & 0.7 & 0.82 & 1.0 \\
 \midrule
\multirow{2}{*}{GPSE} & sparse & 1.0 & 1.0 & 0.25 & 0.95 & 0.52 \\
 & fully-connected & 1.0 & 1.0 & 0.1 & 1.0 & 0.48 \\
 \midrule
\multirow{2}{*}{LapPE} & sparse & 1.0 & 1.0 & 0.75 & 0.65 & 1.0 \\
 & fully-connected & 1.0 & 1.0 & 1.0 & 0.83 & 1.0 \\
 \midrule
\multirow{2}{*}{RWSE} & sparse & 1.0 & 1.0 & 1.0 & 0.97 & 1.0 \\
 & fully-connected & 1.0 & 1.0 & 1.0 & 1.0 & 1.0 \\
  \midrule
\multirow{2}{*}{RRWP} & sparse & 1.0 & 1.0 & 0.5 & 0.65 & 1.0 \\
 & fully-connected & 1.0 & 1.0 & 1.0 & 0.99 & 1.0 \\
 \midrule
\multirow{2}{*}{SignNet} & sparse & 0.75 & 1.0 & 1.0 & 0.89 & 1.0 \\
 & fully-connected & 1.0 & 1.0 & 1.0 & 0.99 & 1.0 \\

\bottomrule
\end{tabular}%
}
    \label{tab:synth_experiments}
\end{table}

We first assess PEs on synthetic datasets designed to test specific structural distinctions. These include \textsc{LIMITS 1} and \textsc{LIMITS 2}~\cite{garg2020generalization}, \textsc{SKIP-CIRCLES}~\cite{chen2019equivalence}, \textsc{TRIANGLES}~\cite{sato2021random}, and \textsc{4-CYCLES}~\cite{loukas2019graph}. We evaluate both sparse and fully connected topologies using GIN and GRIT and report the results in \cref{tab:synth_experiments}.
RWSE consistently achieves perfect scores across tasks and topologies, followed by LapPE and SignNet. GCKN and GPSE show more variable results and perform poorly on structure-sensitive datasets such as \textsc{SKIP-CIRCLES} and \textsc{4-CYCLES}. These results confirm that many PEs are capable of encoding complex structural patterns under idealized conditions.
However, this synthetic expressiveness does not consistently translate to improved performance on real-world datasets. For example, while RWSE achieves perfect accuracy in synthetic tasks, it fails to outperform simpler PEs on benchmarks such as ZINC and Peptides-func. LapPE, in contrast, provides a more balanced profile, performing well in both controlled and practical settings.
These observations highlight the limitations of synthetic evaluations: although useful for isolating specific capabilities, they cannot predict the effectiveness of PEs under more realistic conditions, where noise, feature distributions, and task semantics come into play.

\subsection{Expressiveness $\neq$ Performance}
\begin{figure}[b]
    \centering    \includegraphics[width=\linewidth]{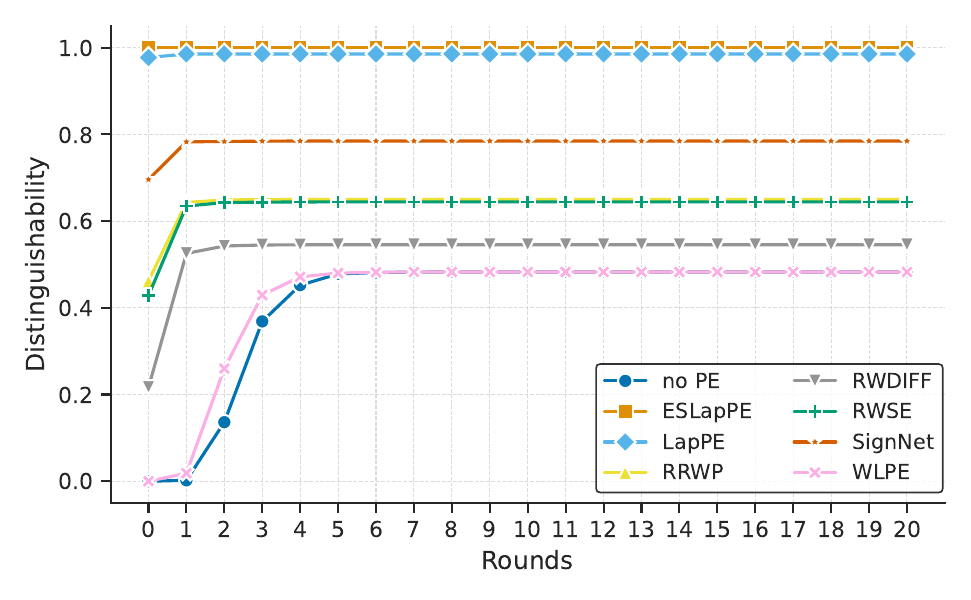}
    \caption{WL distinguishability scores on the ZINC dataset. Spectral PEs such as LapPE and ESLapPE achieve near-perfect node distinguishability. However, this does not consistently align with empirical performance.}
    \label{fig:distinguishability_zinc}
\end{figure}
\begin{figure}
    \centering
    \includegraphics[width=\linewidth]{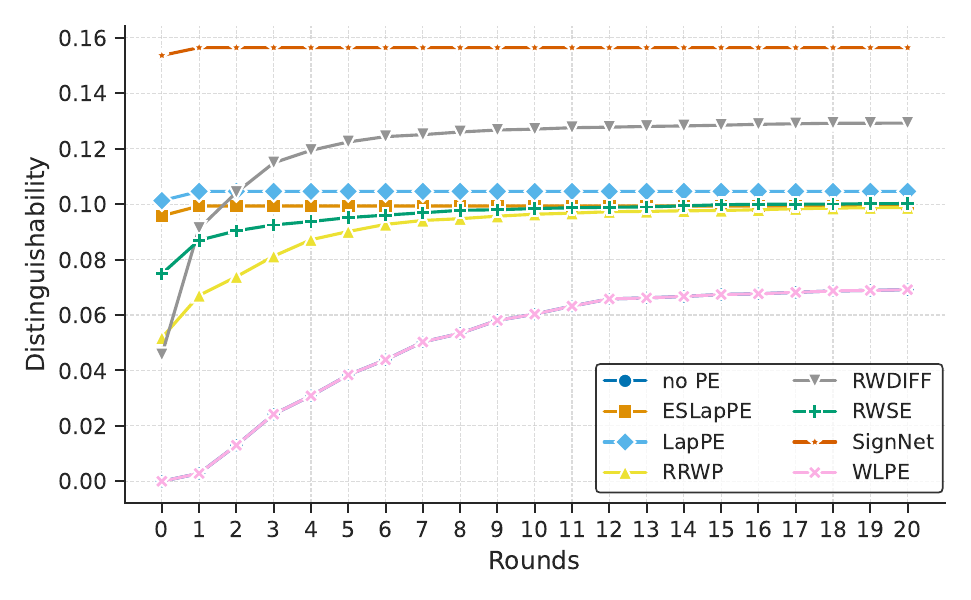}
    \caption{WL distinguishability for Peptides. None of the evaluated configurations achieves full distinguishability and the one with the highest expressiveness (SignNet) is among the weakest performers on this benchmark.}
    \label{fig:distinguishability_peptides}
\end{figure}
\begin{figure}
    \centering
    \includegraphics[width=\linewidth]{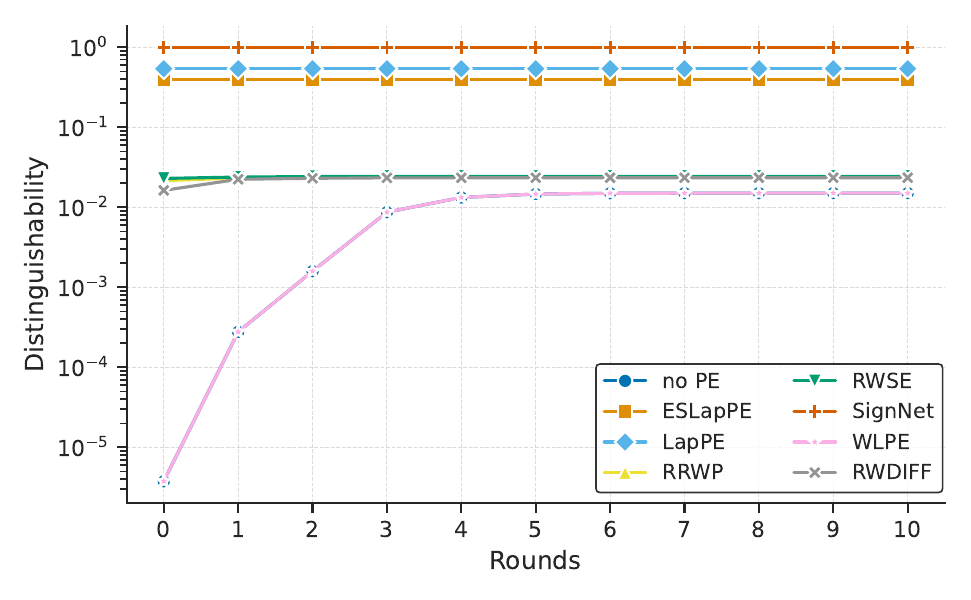}
    \caption{WL distinguishability on PCQM-Contact. Laplacian-based PEs yield much higher expressiveness than random-walk based ones.}
    \label{fig:distinguishability_pcqm}
\end{figure}

To further examine the practical value of theoretical expressiveness, we analyze WL distinguishability induced by different PEs on real-world datasets (formally defined in Section~\ref{sec:wl_distinguishability}). 
Although this metric offers an upper bound on the discriminative capacity that a GNN might achieve with a given PE, it does not imply that full distinguishability is necessary or even beneficial for a given task. In fact, we find that empirical performance often does not align with this score.
For image-derived datasets such as CIFAR10, MNIST, PascalVOC-SP, and COCO-SP, node features (e.g., pixel coordinates) are inherently unique, which renders additional expressiveness from PEs unnecessary. This aligns with our experimental evaluation, where PEs yield marginal improvements in these settings.
More interesting patterns emerge in datasets such as ZINC, Peptides, and PCQM-Contact, which we show in Figure~\ref{fig:distinguishability_zinc}, \ref{fig:distinguishability_peptides} and \ref{fig:distinguishability_pcqm}.
For ZINC, Laplacian-based PEs achieve nearly perfect WL scores but do not translate to the strongest empirical results, which contradicts common expectations. For the Peptides datasets, SignNet exhibits high WL distinguishability but underperforms even the baseline without PE on Peptides-func (\cref{fig:boxplot_lrgb}).
In the PCQM-Contact dataset, we observe a divergence between WL-based expressiveness and empirical performance. Spectral encodings such as LapPE and ESLapPE achieve high WL distinguishability, which indicates a strong theoretical capacity to differentiate node structures. However, similar to our findings on ZINC and Peptides-func, this increased expressiveness does not consistently translate into better predictive performance. In fact, some PEs with lower WL scores, such as RRWP, achieve competitive or even superior results in practice. 

These findings reveal a critical disconnect: higher WL distinguishability does not imply better downstream performance. 
% In some cases, highly expressive PEs may overfit to irrelevant structure or introduce inductive biases that are misaligned with the task. In contrast, tasks grounded in local patterns may benefit little from globally expressive encodings.
One plausible explanation for this disconnect is that highly expressive Positional Encodings capture structural distinctions that are weakly aligned, or even misaligned, with task-relevant signals. In contrast, moderately expressive encodings may provide inductive biases that better match the semantics of real-world prediction tasks, leading to stronger generalization despite lower theoretical expressiveness.
In general, our results show that theoretical expressiveness is not a reliable predictor of practical utility. WL distinguishability, while informative about structural capacity, fails to account for dataset-specific factors such as noise, feature informativeness, and task complexity. Consequently, empirical validation remains essential for PE selection.
These insights emphasize the need for rigorous benchmarking, as outlined in \cref{sec:eval}, to assess the practical effectiveness of PEs. Expressiveness alone does not guarantee better performance, and in some cases may even be detrimental.

\paragraph{Limitations and Scope.}
Our analysis is primarily empirical and descriptive in nature.
While we observe systematic mismatches between WL distinguishability and downstream performance (e.g., SignNet on Peptides or LapPE on ZINC), providing a causal explanation is challenging due to complex interactions between PEs, model architectures, optimization dynamics, and dataset characteristics. Consequently, our study aims to expose  patterns and counterexamples to commonly held assumptions, rather than to make causal claims about the mechanisms underlying PE
effectiveness. 
From a practical perspective, our results suggest that PEs should be selected with care and empirically validated. Encodings that achieve high WL distinguishability do not consistently yield better performance and in some cases may even be detrimental.
In contrast, PEs that offer a more moderate trade-off between expressiveness and computational cost often perform competitively
across datasets. These observations should be interpreted as empirical
regularities within our benchmark, rather than as prescriptive
recommendations.

\section{Conclusions}
\label{sec:conclusions}

This paper presents a comprehensive investigation into the role of Positional Encodings in Graph Neural Networks and Graph Transformers through a systematic and reproducible benchmarking framework. By decoupling PE evaluation from architectural innovations, we assess the empirical utility and theoretical expressiveness of PEs across a broad range of datasets and models.
Our findings reveal that the effectiveness of PEs is highly dependent on the task and the dataset. In particular, theoretical expressiveness, as measured by WL distinguishability, does not reliably predict downstream performance, which underscores the need for empirical validation. While random-walk-based PEs such as RRWP and RWSE frequently perform well, they incur significant preprocessing cost and memory overhead. In our benchmark, spectral PEs (e.g., LapPE, ESLapPE) often provide
a favorable trade-off between accuracy and preprocessing cost. In contrast, image-derived graphs with unique node features (e.g., CIFAR10) sometimes benefit little from PEs, suggesting that their utility depends strongly on input feature distributions.
Our benchmark also provides insights into architectural design. For instance, we show that sparsified attention mechanisms, as implemented in SparseGRIT, can match the performance of fully connected GTs while significantly reducing computational cost. These results challenge the assumption that full attention is always necessary and demonstrate that careful design of local message-passing topologies, combined with suitable PEs, can yield highly competitive models.
To support future research, we publicly release our benchmarking codebase, including standardized implementations of all models and PEs. We hope this resource facilitates more principled evaluation and design of Positional Encodings in graph representation learning.

\newpage

\bibliographystyle{ACM-Reference-Format}
\bibliography{sample-base}

\pagebreak
\clearpage

\appendix

\begin{table*} \scriptsize
\caption{Results for \textsc{BenchmarkingGNNs} including ZINC, MNIST, CIFAR10, PATTERN, and CLUSTER.
}
\label{GNNbenchmark}
\centering 
\resizebox{0.77\textwidth}{!}{%
\begin{tabular}{lccccc}
\toprule 
Sparse Graph &  \textbf{MNIST} $\uparrow$ & \textbf{CIFAR10} $\uparrow$ & \textbf{PATTERN} $\uparrow$ & \textbf{CLUSTER} $\uparrow$ & \textbf{ZINC} $\downarrow$  \\ \midrule
\texttt{GatedGCN + noPE} & $97.800{\pm 0.138}$ & $69.303{\pm 0.318}$ & $85.397{\pm 0.040}$ & $61.695{\pm 0.261}$ & $0.2398{\pm 0.0094}$\\
\texttt{GatedGCN + ESLapPE} & $97.870{\pm 0.090}$ & $69.438{\pm 0.297}$ & $85.422{\pm 0.161}$ & $61.953{\pm 0.082}$ & $0.2409{\pm 0.0131}$\\
\texttt{GatedGCN + LapPE} &$97.575{\pm 0.025}$ & $69.285{\pm 0.205}$& $86.700{\pm 0.000}$& $65.130{\pm 0.405}$ & $0.1718{\pm 0.0024}$\\
\texttt{GatedGCN + RWSE} & $97.840{\pm 0.171}$ & $69.038{\pm 0.152}$ & $86.833{\pm 0.030}$ & $65.675{\pm 0.296}$ & $0.1016{\pm 0.0030}$\\ 
\texttt{GatedGCN + SignNet} & $97.553{\pm 0.167}$  &$68.570{\pm 0.240}$ & $86.763{\pm 0.027}$ & $75.293{\pm 0.047}$ & $0.1060{\pm 0.0021}$\\ 
\texttt{GatedGCN + PPR}  & $97.797{\pm 0.045}$ & $69.224{\pm 0.546}$ & $86.522{\pm 0.093}$ & $74.175{\pm 0.122}$ & $0.3678{\pm 0.0198}$\\ 
\texttt{GatedGCN + GCKN} & $97.745{\pm 0.069}$ & $69.408{\pm 0.222}$ & $86.758{\pm0.049}$ & $62.478{\pm 0.156}$ & $0.1446{\pm 0.0048}$\\ 
\texttt{GatedGCN + WLPE} & $97.693{\pm 0.235}$& $69.418{\pm 0.165}$  & $84.980{\pm 0.160}$ & $62.738{\pm 0.291}$ & $0.1779{\pm 0.0059}$\\ 
\texttt{GatedGCN + RWDIFF}&  $97.823{\pm 0.119}$ &$69.528{\pm 0.494}$ & $86.760{\pm 0.043}$ & $65.653{\pm 0.470}$ & $0.1346{\pm 0.0074}$ \\ 

\texttt{GatedGCN + RRWP} & $97.908{\pm 0.076}$& $69.572{\pm 0.787}$ & $85.465{\pm 0.148}$ & $61.728{\pm 0.174}$ & $0.2451{\pm 0.0131}$ \\ \midrule

\texttt{GINE + noPE} & $97.712{\pm 0.120}$ & $65.554{\pm 0.225}$ & $85.482{\pm 0.272}$ & $48.783{\pm 0.060}$ & $0.1210{\pm 0.0107}$\\
\texttt{GINE + ESLapPE} & $97.596{\pm 0.071}$ & $66.140{\pm 0.310}$ & $85.546{\pm 0.114}$ & $48.708{\pm 0.061}$ & $0.1209{\pm 0.0066}$\\
\texttt{GINE + LapPE} &$97.555{\pm 0.045}$ & $65.325{\pm 0.195}$& $85.835{\pm 0.195}$& $48.685{\pm 0.035}$ & $0.1144{\pm0.0028}$\\
\texttt{GINE + RWSE} & $97.686{\pm 0.073}$& $65.238{\pm 0.283}$ & $86.688{\pm0.084}$ & $50.642{\pm 0.694}$ & $0.0795{\pm 0.0034}$\\ 
\texttt{GINE + SignNet} & $97.692{\pm 0.165}$ & $64.538{\pm 0.314}$ & $86.538{\pm 0.044}$ & $59.660{\pm 0.630}$ & $0.0993{\pm 0.0069}$ \\ 
\texttt{GINE + PPR}  & $97.650{\pm 0.088}$ & $65.082{\pm 0.434}$ & $85.658{\pm 0.048}$ & $47.440{\pm 2.290}$ & $0.3019{\pm 0.0122}$\\ 
\texttt{GINE + GCKN} & $97.708{\pm 0.105}$ & $65.976{\pm 0.308}$ & $85.844{\pm 0.157}$ & $48.780{\pm 0.149}$ & $0.1169{\pm 0.0029}$\\ 
\texttt{GINE + WLPE} & $97.716{\pm 0.118}$ & $66.132{\pm 0.225}$ & $85.676{\pm 0.084}$  & $48.997{\pm 0.068}$ & $0.1205{\pm 0.0062}$\\ 
\texttt{GINE + RWDIFF}& $97.750{\pm 0.097}$ & $65.632{\pm 0.553}$ & $85.764{\pm 0.209}$ & $49.148{\pm 0.168}$ &$0.0750{\pm 0.0058}$ \\

\texttt{GINE + RRWP} & $96.742{\pm 0.277}$ & $62.790{\pm 1.501}$ & $86.526{\pm 0.036}$ & $48.736{\pm 0.108}$ & $0.0857{\pm 0.0009}$\\

\midrule
\texttt{Exphormer + noPE} & $98.414{\pm 0.047}$ & $74.962{\pm 0.631}$ & $85.676{\pm 0.049}$ & $77.500{\pm 0.151}$ & $0.1825{\pm 0.0209}$ \\
\texttt{Exphormer + ESLapPE}    & $98.354{\pm 0.108}$ &    $74.880{\pm 0.322}$    &  $86.734{\pm 0.024}$  & $78.218{\pm 0.267}$  & $0.2023{\pm 0.0140}$ \\ 
\texttt{Exphormer + LapPE}  & $98.270{\pm 0.070}$ & $75.205{\pm 0.095}$& $86.565{\pm 0.075}$ & $77.175{\pm 0.165}$& $0.1503{\pm 0.0117}$ \\ 
\texttt{Exphormer + RWSE}    & $98.254{\pm 0.084}$    &   $74.434 {\pm 0.205}$   &  $86.820{\pm 0.040}$ &$ 77.690{\pm 0.147}$ & $0.0933{\pm 0.0050}$   \\ 
\texttt{Exphormer + SignNet} &  $98.136{\pm 0.094}$    &  $73.842 {\pm 0.317}$     & $86.752{\pm 0.088}$ & $78.280{\pm  0.211}$  & $0.0924{\pm 0.0072}$ \\ 
\texttt{Exphormer + PPR}    & $98.076{\pm 0.126}$      &   $74.076 {\pm 0.104}$   &  $86.712{\pm 0.047}$ &  $78.098{\pm 0.211}$  & $0.2414{\pm 0.0123}$\\ 
\texttt{Exphormer + GCKN}           & $98.402{\pm 0.067}$    &   $74.926{\pm 0.288}$   & $86.730{\pm 0.040}$  &  $77.470{\pm  0.067}$ & $0.1690{\pm 0.0056}$ \\ 
\texttt{Exphormer + WLPE}  &  $98.398{\pm 0.162}$      &   $74.794{\pm 0.358}$   & $85.454{\pm 0.033}$ &  $77.402{\pm  0.120}$ & $0.1465{\pm 0.0095}$\\ 
\texttt{Exphormer + RWDIFF} &   $98.416{\pm 0.055}$   &   $74.886{\pm 0.810}$  & $86.792{\pm 0.023}$ & $ 77.550{\pm 0.057}$ & $0.1360{\pm 0.0082}$\\
\texttt{Exphormer + RRWP}  &   $98.418{\pm 0.179}$      & $74.504{\pm 0.369}$       &  $85.652{\pm 0.001}$ &  $77.434{\pm 0.056}$ & $0.1914{\pm 0.0153}$\\

\midrule
\texttt{GraphGPS + noPE} & $98.136{\pm 0.085}$ & $72.310{\pm 0.198}$ & $84.182{\pm 0.276}$ & $77.590{\pm 0.158}$ & $0.1610{\pm 0.0045}$\\
\texttt{GraphGPS + ESLapPE}   &  $98.180{\pm  0.117}$       &   $72.122{\pm   0.511}$       &  $86.700{\pm 0.055}$ & $77.800 {\pm0.107}$ & $0.1795{\pm 0.0110}$\\ 
\texttt{GraphGPS + LapPE}   & $98.065{\pm 0.075}$&  $72.310{\pm 0.530}$ & $86.550{\pm 0.150}$&  $77.355{\pm 0.115}$  & $0.1086{\pm0.0062}$\\ 
\texttt{GraphGPS + RWSE}   &   $98.116{\pm 0.102} $     &    $ 72.034{\pm 0.756 }$   & $86.866{\pm 0.010}$ &  $77.550 {\pm 0.195}$ & $0.0744{\pm 0.0060}$  \\ 
\texttt{GraphGPS + SignNet} &  $98.012{\pm 0.091}$      &    $72.152{\pm  0.323}$    & $86.734{\pm 0.069}$ & $78.308{\pm 0.111}$ & $0.0945{\pm 0.0019}$  \\ 
\texttt{GraphGPS + PPR}    & $98.010{\pm 0.097}$        &    $ 71.842{\pm 0.325}$    & $86.124{\pm 0.214}$ &  $76.828 {\pm 0.250}$ & $0.1349{\pm 0.0054}$\\ 
\texttt{GraphGPS + GCKN}  &  $98.180{\pm 0.117}$     &   $72.194{\pm  0.515 }$     & $86.786{\pm 0.043}$& $77.514 {\pm 0.182}$ &  $0.1460{\pm 0.0078}$ \\ 
\texttt{GraphGPS + WLPE}  &  $98.038{\pm 0.134}$       &   $72.258  {\pm 0.661}$    & $84.916{\pm 0.195}$ & $76.866 {\pm 0.171}$ & $0.1204{\pm 0.0055}$\\ 
\texttt{GraphGPS + RWDIFF} &   $98.026{\pm 0.101}$     &   $71.800{\pm 0.363}$    & $86.820{\pm 0.063}$  &  $77.478 {\pm0.150}$ & $0.0924{\pm 0.0212}$ \\ 
\texttt{GraphGPS + RRWP}  &  $98.146{\pm 0.105}$       &    $72.084{\pm 0.466}$    & $84.436{\pm 0.224}$ &  $77.420{\pm 0.080}$ & $0.1690{\pm 0.0084}$\\

\midrule
 \texttt{SparseGRIT + noPE} & $97.940{\pm 0.071}$ & $72.778{\pm 0.627}$ & $85.948{\pm 0.148}$ & $77.274{\pm 0.170}$ & $0.1255{\pm 0.0062}$\\
\texttt{SparseGRIT + ESLapPE}    & $97.970{\pm 0.110}$     &  $72.494{\pm 0.501}$    & $86.018{\pm 0.319}$ &  $77.238  {\pm 0.066}$ & $0.1280{\pm 0.0077}$\\ 
\texttt{SparseGRIT + LapPE}    & $97.915{\pm 0.065}$  & $72.640{\pm 0.040}$ & $86.555{\pm 0.025}$& $76.100{\pm 0.085}$& $0.1070{\pm 0.0017}$\\ 
\texttt{SparseGRIT  + RWSE}    &   $98.122{\pm 0.054}$     &   $ 72.330 {\pm 0.600}$    & $86.914{\pm 0.031}$  & $77.148  {\pm 0.174}$ & $0.0676{\pm 0.0060}$\\ 
\texttt{SparseGRIT  + SignNet} &  $97.946{\pm 0.122}$      &   $71.003 {\pm 0.301}$   & $86.794{\pm 0.055}$ &  $78.882  {\pm  0.146}$ & $0.0821{\pm 0.0043}$\\ 
\texttt{SparseGRIT + PPR}    & $98.020{\pm 0.194}$       &    $71.926 {\pm 0.833}$      & $86.650{\pm 0.033}$ &  $78.732 {\pm 0.202}$ & $0.2536{\pm 0.0193}$\\ 
\texttt{SparseGRIT  + GCKN }  & $97.958{\pm 0.127}$       &  $72.598 {\pm 0.535 }$    & $86.650{\pm 0.033}$  &  $76.746 {\pm  0.187}$ & $0.1233{\pm 0.0071}$\\ 
\texttt{SparseGRIT  + WLPE}  & $97.946{\pm 0.125}$       &  $ 72.096{\pm 0.835 }$     & $85.712{\pm 0.027}$ &  $77.170 {\pm0.143}$ & $0.1262{\pm 0.0059}$\\ 
\texttt{SparseGRIT  + RWDIFF} & $98.022{\pm 0.083}$       &   $72.366 {\pm 0.388 }$    & $86.938{\pm 0.045}$  &  $77.214  {\pm 0.065}$ & $0.0690{\pm 0.0039}$\\ 
\texttt{SparseGRIT  + RRWP } & $98.088{\pm 0.048}$   &  $74.954{\pm 0.256}$      & $87.168{\pm 0.041}$ & $79.872{\pm 0.079}$ & $0.0651{\pm 0.0027}$\\

\midrule
\texttt{GRIT + noPE}    &   $98.108{\pm 0.190}$     &   $74.402{\pm 0.135}$ & $87.126{\pm 0.033}$ &  $78.616{\pm 0.178 }$& $0.1237{\pm 0.0057}$ \\        
\texttt{GRIT + ESLapPE}    &  $98.010{\pm 0.141}$      &  $74.558{\pm 0.682}$   & $87.140{\pm 0.064} $ & $78.588{\pm 0.111}$ & $0.1241{\pm0.0031}$  \\ 
\texttt{GRIT + LapPE}    &  $97.875{\pm 0.001}$ & $73.325{\pm 0.505}$ & $86.985{\pm 0.015}$ & $77.960{\pm 0.310}$ & $0.1039{\pm 0.0035}$\\ 
\texttt{GRIT + RWSE}    &   $98.068{\pm 0.182}$     &  $73.652{\pm 0.623}$      & $87.116{\pm 0.046}$ & $78.880{\pm0.057}$ & $0.0671{\pm 0.0037}$\\ 
\texttt{GRIT + SignNet} &   $97.766{\pm 0.220}$     &  $72.812{\pm 0.482}$     &  $87.085{\pm 0.064}$ & $79.770{\pm 0.150} $ & $0.0945{\pm 0.0098}$\\ 
\texttt{GRIT + PPR}    &  $97.986{\pm 0.082}$     &  $73.568{\pm 0.451}$    &  $86.780{\pm 0.001}$ & $78.958{\pm 0.175}$ & $0.1390{\pm 0.0076}$ \\ 
\texttt{GRIT + GCKN}   &  $98.084{\pm 0.139}$        &  $73.946{\pm 0.910}$    &  $87.194{\pm 0.044}$        &   $78.542{\pm 0.149}$ & $0.1306{\pm 0.0141}$\\ 
\texttt{GRIT + WLPE}  &  $98.022{\pm 0.173}$        & $74.206{\pm 0.684}$     & $86.863{\pm 0.033}$  & $78.500{\pm 0.091}$  & $0.1218{\pm 0.0035}$\\ 
\texttt{GRIT + RWDIFF} & $98.024{\pm 0.148}$       &  $73.956{\pm 0.202}$   &  $87.152{\pm 0.045}$&  $78.778{\pm 0.090}$ & $0.0671{\pm 0.0060}$\\ 
\texttt{GRIT + RRWP}  &  $98.124{\pm 0.141}$      &   $75.662{\pm 0.410}$
&  $87.217{\pm 0.034}$& $79.812{\pm 0.109}$ & $0.0590{\pm 0.0010}$\\  \bottomrule
\end{tabular}
}
\end{table*}

\begin{table*} \scriptsize
\caption{Results for LRGB datasets including Peptides\_func,  Peptides\_struct, PCQM\_Contact, PascalVOC-SuperPixels and COCO-SuperPixels.
}
\centering
\label{LRGB}
\resizebox{\textwidth}{!}{%
\begin{tabular}{lccccc}
\toprule
  Sparse Graph           & Peptides-func &  Peptides-struct & PCQM-Contact & PascalVOC-SP  & COCO-SP \\ \midrule
  \texttt{GatedGCN + noPE} & $0.6523{\pm 0.0074}$& $0.2470{\pm 0.0005}$ & $0.4730{\pm 0.0003}$& $0.3923{\pm 0.0020}$ & $0.2619{\pm 0.0045}$ \\
\texttt{GatedGCN + LapPE}&  $0.6581{\pm 0.0068}$& $0.2472{\pm 0.0003}$&$0.4764{\pm 0.0004}$ & $0.3920{\pm 0.0033}$ & $0.2671{\pm 0.0006}$\\
\texttt{GatedGCN + ESLapPE}& $0.6484{\pm 0.0037}$ & $0.2490{\pm 0.0020}$ & $0.4736{\pm 0.0006}$ & $0.3930{\pm 0.0041}$ & $0.2628{\pm 0.0004}$\\
\texttt{GatedGCN + RWSE} & $0.6696{\pm 0.0022}$ & $0.2485{\pm 0.0022}$&$0.4749{\pm 0.0005}$ & $0.3882{\pm 0.0041}$ &$0.2657{\pm 0.0007}$ \\
\texttt{GatedGCN + SignNet} & $0.5327{\pm 0.0137}$ & $0.2688{\pm 0.0016}$ & $0.4672{\pm 0.0001}$ &$0.3814{\pm 0.0005}$ & -\\
\texttt{GatedGCN + GCKN} &  $0.6544{\pm 0.0040}$& 
 $0.2483{\pm 0.0009}$  & $0.4687{\pm 0.0002}$ &   $0.3933{\pm 0.0044}$ & - \\
\texttt{GatedGCN + WLPE} & $0.6562{\pm 0.0053}$ &  $0.2473{\pm 0.0012}$& $0.4671{\pm 0.0003}$ &$0.3805{\pm 0.0018}$ & - \\
\texttt{GatedGCN + RWDIFF} & $0.6527{\pm 0.0053}$& $0.2474{\pm 0.0003}$& $0.4740{\pm 0.0003}$ &$0.3919{\pm 0.0019}$ & $0.2674{\pm 0.0031}$ \\
\texttt{GatedGCN + RRWP} &$0.6516{\pm 0.0072}$ & $0.2514{\pm 0.0001}$& - & - & - \\
\midrule
\texttt{GraphGPS + noPE} & $0.6514{\pm 0.0123}$ & $0.4243{\pm 0.0305}$ & $0.4649{\pm 0.0025}$& $0.4517{\pm 0.0112}$ & $0.3799{\pm 0.0056}$\\            
\texttt{GraphGPS + LapPE} & $0.6620{\pm 0.0073}$ & $0.2497{\pm 0.0024}$ & $0.4696{\pm 0.0017}$ & $0.4505{\pm 0.0062 }$ & $0.3859{\pm 0.0016}$ \\ 
\texttt{GraphGPS + ESLapPE}    & $0.6516{\pm 0.0062}$     &  $0.2568{\pm 0.0013}$  & $0.4639{\pm 0.0031}$ & $0.4538{\pm 0.0083}$ & $0.3866{\pm 0.0017}$ \\ 
\texttt{GraphGPS + RWSE}    &  $0.6510{\pm 0.0071}$       &  $0.2549{\pm 0.0033}$  & $0.4685{\pm 0.0009}$& $0.4531{\pm 0.0073}$ & $0.3891{\pm 0.0033}$ \\ 
\texttt{GraphGPS + SignNet} &   $0.5719 {\pm  0.0055}$   &  $0.2657{\pm 0.0021}$    & $0.4624{\pm 0.0020}$ & $0.4291{\pm 0.0056}$ & -   \\ 
\texttt{GraphGPS + GCKN}  &    $0.6502{\pm 0.0101 }$     &    $0.2519{\pm   0.0005 }$            & $0.4609{\pm 0.0007}$& $0.4515 {\pm 0.0053}$ & - \\ 
\texttt{GraphGPS + WLPE}  &    $0.5851{\pm 0.0441}$ &  $0.5203{\pm 0.0504}$   & $0.4622{\pm 0.0012}$& $0.4501{\pm 0.0057}$ & -\\ 
\texttt{GraphGPS + RWDIFF} & $0.6519 {\pm 0.0077}$       & $0.4769{\pm 0.0360 }$   & $0.4669{\pm 0.0006}$& $0.4488{\pm 0.0097}$ & $0.3873{\pm 0.0024}$\\ 
\texttt{GraphGPS + RRWP}  &   $0.6505{\pm 0.0058}$     &  $0.3734{\pm 0.0157}$    &  - & -  & -\\ 
\midrule
\texttt{Exphormer + noPE} & $0.6200{\pm 0.0052}$ & $0.2584{\pm 0.0019}$ & $0.4661{\pm 0.0021}$ & $0.4149{\pm 0.0047}$ & $0.3445{\pm 0.0052}$\\
\texttt{Exphormer + LapPE} & $0.6424{\pm 0.0063}$ & $0.2496{\pm 0.0013}$ & $0.4737{\pm 0.0024 }$ & $0.4242{\pm 0.0044}$ & $0.3471{\pm 0.0028}$\\
\texttt{Exphormer + ESLapPE} & $0.6281{\pm 0.0085}$ & $0.2513{\pm 0.0022}$ & $0.4676{\pm 0.0018}$ & $0.4141{\pm 0.0054}$ & $0.3485{\pm 0.0011}$ \\
\texttt{Exphormer + RWSE} & $0.6240{\pm 0.0069}$ & $0.2579{\pm 0.0010}$ & $0.4642{\pm 0.0039}$  & $0.4218{\pm 0.0063}$ & $0.3485{\pm 0.0011}$\\
\texttt{Exphormer + SignNet} & $0.5458{\pm 0.0097}$ & $0.2667{\pm 0.0037}$ & $0.4615{\pm 0.0066}$ & $0.3966{\pm 0.0020}$ & -\\
\texttt{Exphormer + GCKN} & $0.6422{\pm 0.0080}$ & $0.2514{\pm 0.0012}$ & $0.4604{\pm 0.0038}$ & $0.4196{\pm 0.0049}$ & -\\
\texttt{Exphormer + WLPE} & $0.6216{\pm 0.0069}$ & $0.2558{\pm 0.0011}$  & $0.2051{\pm 0.0080 }$ & $0.4104{\pm 0.0071}$ & - \\
\texttt{Exphormer + RWDIFF} & $0.6275{\pm 0.0031}$ & $0.2556{\pm 0.0021}$ & $0.4642{\pm 0.0032}$& $0.4165{\pm 0.0059}$ & $0.3417{\pm 0.0006}$\\
\texttt{Exphormer + RRWP} & $0.6208{\pm 0.0074}$ & $0.2586{\pm 0.0014}$  & - & - & -\\ 

\midrule
\texttt{SparseGRIT + noPE}  & $0.4885{\pm 0.0036}$& $0.2550{\pm 0.0006}$ & $0.4527{\pm 0.0006}$ & $0.3471{\pm 0.0030}$ & $0.1976{\pm 0.0038}$\\
\texttt{SparseGRIT + LapPE} & $0.5884{\pm 0.0059}$&$0.2487{\pm 0.0014}$ & $0.4585{\pm 0.0011}$ & $0.3514{\pm 0.0026}$ & $0.1974{\pm 0.0008}$\\
\texttt{SparseGRIT + ESLapPE}& $0.5161{\pm 0.0069}$&$0.2537{\pm 0.0005}$ & $0.4532{\pm 0.0005}$& $0.3462{\pm 0.0035}$ & $0.1958{\pm 0.0001}$\\
 \texttt{SparseGRIT + RWSE} & $0.5570{\pm 0.0079}$ & $0.2537{\pm 0.0012}$ & $0.4553{\pm 0.0014}$ & $0.3460{\pm 0.0071}$ & $0.1969{\pm 0.0010}$\\
\texttt{SparseGRIT + SignNet} & $0.5115{\pm 0.0064}$&$0.2640{\pm 0.0018}$ &$0.4573{\pm 0.0003}$ & $0.3419{\pm 0.0074}$ & - \\
\texttt{SparseGRIT + GCKN} & $0.5871{\pm 0.0042}$&$0.2492{\pm 0.0010}$ & $ 0.4500{\pm 0.0004}$& $0.3519{\pm 0.0040}$ & -\\
\texttt{SparseGRIT + WLPE} & $0.4808{\pm 0.0016}$ &  $0.2547{\pm 0.0005}$ & $0.4489{\pm 0.0012}$& $0.3439{\pm 0.0027}$ & -\\
\texttt{SparseGRIT + RWDIFF} & $0.5521{\pm 0.0072}$ & $0.2550{\pm 0.0008}$ & $0.4551{\pm 0.0005}$ & $0.3447{\pm 0.0046}$ & $0.1965{\pm 0.0011}$\\
\texttt{SparseGRIT + RRWP} &  $0.6702{\pm 0.0080}$ & $0.2504{\pm 0.0025}$& - & - & -\\ 
\midrule
\texttt{GRIT + noPE} & $0.4861{\pm 0.0053}$ & $0.2489{\pm 0.0008}$ & $0.4525 \pm 0.0001$ & $0.3556 \pm 0.0019$ & $0.2105 \pm 0.0004$ \\
\texttt{GRIT + LapPE}&  $0.5834{\pm 0.0105}$ & $0.2474{\pm 0.0005}$ & $0.4580 \pm 0.0020$ & $0.3551 \pm 0.0032$ & $0.2112 \pm 0.0005$ \\
\texttt{GRIT + ESLapPE}& $0.4831{\pm 0.0023}$ & $0.2584{\pm 0.0002}$ & $0.4486 \pm 0.0014$ & $0.3485 \pm 0.0028$ & $0.2100 \pm 0.0008$ \\
\texttt{GRIT + RWSE} & $0.5432{\pm 0.0034}$ & $0.2612{\pm 0.0008}$ & $0.4524 \pm 0.0001$ & $0.3461 \pm 0.0058$ & $0.2114 \pm 0.0009$ \\
\texttt{GRIT + SignNet} & $0.5307{\pm 0.0085}$ & $0.2600{\pm 0.0018}$ & $0.4608 \pm 0.0007$ & $0.3385 \pm 0.0045$ & - \\
\texttt{GRIT + GCKN} & $0.5868{\pm 0.0051}$ & $0.2477{\pm 0.0006}$ & $0.4521 \pm 0.0002$ & $0.3516 \pm 0.0003$ & - \\
\texttt{GRIT + WLPE} & $0.4798{\pm 0.0012}$ & $0.2578{\pm 0.0011}$ & $0.4515 \pm 0.0004$ & $0.3441 \pm 0.0011$ & - \\
\texttt{GRIT + RWDIFF} & $0.5801{\pm 0.0036}$ & $0.2639{\pm 0.0010}$ & $0.4563 \pm 0.0003$ & $0.3521 \pm 0.0079$ & $0.2128 \pm 0.0008$ \\
\texttt{GRIT + RRWP} & $0.6865{\pm 0.0050}$ & $0.2454{\pm 0.0010}$ & - & - & - \\
\bottomrule
\end{tabular}
}
\end{table*}

\end{document}